\documentclass[]{interact}

\usepackage{epstopdf}
\usepackage{subfigure}
\usepackage{tabularx}
\usepackage{graphicx}
\usepackage{amsmath}
\usepackage{amssymb}
\usepackage{booktabs}
\usepackage{placeins} 
\usepackage{float} 
\usepackage{enumitem}
\usepackage{tabularx} 
\usepackage{longtable} 
\usepackage{array} 
\usepackage{ragged2e} 
\usepackage{booktabs} 
\usepackage{placeins}
\usepackage{afterpage} 
\usepackage{caption} 
\usepackage{booktabs} 
\usepackage{adjustbox} 
\usepackage{graphicx} 
\usepackage{tabularx}

\theoremstyle{plain}

\theoremstyle{definition}

\theoremstyle{remark}

\begin{document}


\title{Expert Kernel Generation Network Driven by Contextual Mapping for Hyperspectral Image Classification}

\author{
\name{Guandong Li\textsuperscript{a}\thanks{CONTACT Guandong Li. Email: leeguandon@gmail.com} and Mengxia Ye\textsuperscript{b}}
\affil{\textsuperscript{a}iFLYTEK, Shushan, Hefei, Anhui, China; \textsuperscript{b}Aegon THTF,Qinghuai,Nanjing,Jiangsu,China}
}

\maketitle

\begin{abstract}
Deep neural networks face several challenges in hyperspectral image classification, including high-dimensional data, sparse distribution of ground objects, and spectral redundancy, which often lead to classification overfitting and limited generalization capability. To more efficiently adapt to ground object distributions while extracting image features without introducing excessive parameters and skipping redundant information, this paper proposes EKGNet based on an improved 3D-DenseNet model, consisting of a context-aware mapping network and a dynamic kernel generation module. The context-aware mapping module translates global contextual information of hyperspectral inputs into instructions for combining base convolutional kernels, while the dynamic kernels are composed of K groups of base convolutions, analogous to K different types of experts specializing in fundamental patterns across various dimensions. The mapping module and dynamic kernel generation mechanism form a tightly coupled system - the former generates meaningful combination weights based on inputs, while the latter constructs an adaptive expert convolution system using these weights. This dynamic approach enables the model to focus more flexibly on key spatial structures when processing different regions, rather than relying on the fixed receptive field of a single static convolutional kernel. EKGNet enhances model representation capability through a 3D dynamic expert convolution system without increasing network depth or width. The proposed method demonstrates superior performance on IN, UP, and KSC datasets, outperforming mainstream hyperspectral image classification approaches.
\end{abstract}


\section{Introduction}

Hyperspectral remote sensing images (HSI) play a crucial role in spatial information applications due to their unique narrow-band imaging characteristics. The imaging equipment synchronously records both spectral and spatial position information of sampling points, integrating them into a three-dimensional data structure containing two-dimensional space and one-dimensional spectrum. As an important application of remote sensing technology, ground object classification demonstrates broad value in fields including ecological assessment, transportation planning, agricultural monitoring, land management, and geological surveys \cite{bing2011intelligent,chang2003hyperspectral}. However, hyperspectral remote sensing images face several challenges in ground object classification. First, hyperspectral data typically exhibits high-dimensional characteristics, with each pixel containing hundreds or even thousands of bands, leading to data redundancy and computational complexity while potentially causing the "curse of dimensionality," making classification models prone to overfitting under sparse sample conditions. Second, sparse ground object distribution means training samples are often limited, particularly for certain rare categories where annotation costs are high and distributions are imbalanced, further constraining model generalization capability. Additionally, hyperspectral images are frequently affected by noise, atmospheric interference, and mixed pixels, reducing signal-to-noise ratio and increasing difficulty in extracting features from sparse ground objects. Finally, ground object sparsity may also lead to insufficient spatial context information, causing loss in classification and recognition.

Deep learning methods for HSI classification \cite{li2019doubleconvpool,li2020hyperspectral,li2022faster,li2023dgcnet,li2025spatialgeometry,li20253dwaveletconvolutionsextended} have achieved significant progress. In \cite{lee2017going} and \cite{zhao2016spectral}, Principal Component Analysis (PCA) was first applied to reduce the dimensionality of the entire hyperspectral data, followed by extracting spatial information from neighboring regions using 2D CNN. Methods like 2D-CNN \cite{makantasis2015deep,chen2016deep} require separate extraction of spatial and spectral features, failing to fully utilize joint spatial-spectral information and necessitating complex preprocessing. \cite{wang2018fast} proposed a Fast Dense Spectral-Spatial Convolutional Network (FDSSC) based on dense networks, constructing 1D-CNN and 3D-CNN dense blocks connected in series. FSKNet \cite{li2022faster} introduced a 3D-to-2D module and selective kernel mechanism, while 3D-SE-DenseNet \cite{li2020hyperspectral} incorporated the SE mechanism into 3D-CNN to correlate feature maps between different channels, activating effective information while suppressing ineffective information in feature maps. DGCNet \cite{li2023dgcnet} designed dynamic grouped convolution (DGC) on 3D convolution kernels, where DGC introduces small feature selectors for each group to dynamically determine which part of input channels to connect based on activations of all input channels. Multiple groups can capture different complementary visual/semantic features of input images, enabling CNNs to learn rich feature representations. DHSNet \cite{liu2025dual} proposed a novel Central Feature Attention-Aware Convolution (CFAAC) module that guides attention to focus on central features crucial for capturing cross-scene invariant information. To leverage the advantages of both CNN and Transformer, many studies have attempted to combine CNN and Transformer to utilize local and global feature information of HSI. \cite{sun2022spectral} proposed a Spectral-Spatial Feature Tokenization Transformer (SSFTT) network that extracts shallow features through 3D and 2D convolutional layers and uses Gaussian-weighted feature tokens to extract high-level semantic features in the transformer encoder. Some Transformer-based methods \cite{hong2021spectralformer,zhao2016spectral,liu2021swin} employ grouped spectral embedding and transformer encoder modules to model spectral representations, but these methods have obvious shortcomings - they treat spectral bands or spatial patches as tokens and encode all tokens, resulting in significant redundant computations. However, HSI data already contains substantial redundant information, and their accuracy often falls short compared to 3D-CNN-based methods while requiring greater computational complexity.

3D-CNN possesses the capability to sample simultaneously in both spatial and spectral dimensions, maintaining the spatial feature extraction ability of 2D convolution while ensuring effective spectral feature extraction. 3D-CNN can directly process high-dimensional data, eliminating the need for preliminary dimensionality reduction of hyperspectral images. However, the 3D-CNN paradigm has significant limitations - when simultaneously extracting spatial and spectral features, it may incorporate irrelevant or inefficient spatial-spectral combinations into computations. For instance, certain spatial features may be prominent in specific bands while appearing as noise or irrelevant information in other bands, yet 3D convolution still forcibly combines these low-value features. Computational and dimensional redundancy can easily trigger overfitting risks, further limiting model generalization capability. Currently widely used methods such as DFAN \cite{zhang2020deep}, MSDN \cite{zhang2019multi}, 3D-DenseNet \cite{zhang2019three}, and 3D-SE-DenseNet\cite{li2020hyperspectral} employ operations like dense connections. While dense connections directly link each layer to all its preceding layers, enabling feature reuse, they introduce redundancy when subsequent layers do not require early features. Therefore, how to more efficiently enhance the representational capability of 3D convolution kernels in 3D convolution, achieving more effective feature extraction with fewer cascaded 3D convolution kernels and dense connections while optimizing the screening and skipping of redundant information has become a direction in hyperspectral classification.

This paper proposes EKGNet, whose core is a context-aware mapping network represented as function $ f_{\text{map}} $. It acts as an intelligent translator: first extracting global context summary $ g = \text{AvgPool}(X) $ from input feature map $ X $, then $ f_{\text{map}} $ translates this context $ g $ into a set of input-specific instructions - attention weights $ \alpha \in \mathbb{R}^K $. Specifically, $ \alpha = \text{Softmax}(f_{\text{map}}(g)/\tau) $, where $ K $ is the number of base kernels and $ \tau $ is a temperature parameter controlling the sharpness of attention distribution. These dynamically generated attention weights $ \alpha = \{\alpha_1,...,\alpha_K\} $ then guide the operation of an expert convolution system containing $ K $ base convolutional kernels $ \{W_1,...,W_K\} $ learned during training, which can be viewed as "experts" each specializing in capturing specific patterns. The final dynamic convolutional kernel $ W_{\text{dyn}} $ is generated by weighted aggregation of these base kernels: $ W_{\text{dyn}} = \sum_{k=1}^K \alpha_k W_k $. Essentially, the context-aware mapping network ($ f_{\text{map}} $) and expert convolution system (implemented through $ W_{\text{dyn}} = \sum \alpha_k W_k $) form a tightly coupled, synergistic whole: the former generates meaningful combination weights ($ \alpha $) based on deep understanding of inputs, while the latter dynamically "assembles" a highly adaptive convolution operation using these weights (e.g., output $ Y = W_{\text{dyn}} * X' $, where $ X' $ is the corresponding input feature). We believe this mechanism combining global context understanding with dynamic expert knowledge can significantly improve the network's adaptability to complex variations in HSI data, achieving breakthroughs in feature extraction and ultimately obtaining superior classification performance and stronger model robustness. This dynamic approach enables the model to focus more flexibly on key spatial structures when processing different spatial regions, rather than relying on the fixed receptive field of a single static convolutional kernel. Compared to traditional 3D-CNN, this method enhances spatial feature representation capability through attention mechanisms without increasing network depth or width, especially when targets in hyperspectral images are unevenly distributed or have complex spatial patterns, enabling more efficient extraction of local features. The high-dimensional spectral information provides fine-grained discrimination basis for classification tasks. EKGNet can reduce aggregation of redundant spectral information, avoiding the uniform sampling and dimensionality reduction treatment of all spectral dimensions in traditional 3D-CNN, effectively alleviating parameter redundancy and computational complexity caused by excessive spectral dimensions.

The main contributions of this paper are as follows:

1. This paper proposes a context-aware mapping network and dynamically varying expert convolution system, improving the spatial-spectral joint hyperspectral image classification method based on efficient 3D-DenseNet. EKGNet dynamically generates convolutional kernels to address overfitting risks caused by sparse ground objects and information redundancy in hyperspectral data, enhancing network generalization capability. By combining dense connections with dynamic convolution generation - where dense connections facilitate feature reuse in the network - the designed 3D-DenseNet model achieves good accuracy on both IN and UP datasets.

2. This paper introduces a context-aware mapping module in 3D-CNN that translates global contextual information of hyperspectral inputs into instructions for combining base convolutional kernels, while dynamic kernels are composed of K groups of base convolutions, analogous to K different types of experts specializing in fundamental patterns across various dimensions, achieving more efficient feature extraction capability.

3. EKGNet is more concise than networks combining various DL mechanisms, without complex connections and concatenations, requiring less computation. Without increasing network depth or width, it improves model representation capability through wavelet convolution with expanded receptive fields.

\section{Context-Aware Mapping Driven Expert Convolution System}

\subsection{Dynamic Convolution}

Given the sparse and finely clustered characteristics of hyperspectral ground objects, most methods consider how to enhance the model's feature extraction capability regarding spatial-spectral dimensions, making dynamic convolution an excellent approach to strengthen spatial-spectral dimension information representation at the convolutional kernel level. LGCNet\cite{li2025spatial} designed a learnable grouped convolution structure where both input channels and convolution kernel groups can be learned end-to-end through the network. DGCNet\cite{li2023dgcnet} designed dynamic grouped convolution, introducing small feature selectors for each group to dynamically determine which part of input channels to connect based on activations of all input channels. DACNet\cite{li2025efficient} designed dynamic attention convolution using SE to generate weights and multiple parallel convolutional kernels instead of single convolution. SG-DSCNet\cite{li2025spatialgeometry} designed a Spatial-Geometry Enhanced 3D Dynamic Snake Convolutional Neural Network, introducing deformable offsets in 3D convolution to increase kernel flexibility through constrained self-learning processes, thereby enhancing the network's regional perception of ground objects and proposing multiview feature fusion. These methods are all based on the design concept of dynamic convolution, attempting to address the complexity of hyperspectral data in spatial-spectral dimensions through adaptive convolution mechanisms. For the information redundancy introduced by 3D-CNN in spatial-spectral dimensions, this paper designs a more reasonable context-aware mapping network and corresponding expert convolution system for dynamic kernel generation from the fundamental characteristics of hyperspectral data, fundamentally solving the problem of ground object recognition loss by capturing global hyperspectral features and generating convolutional weights that fit hyperspectral data.

\subsection{Context-Aware Mapping Module and Expert Convolution System}

Hyperspectral image sample data is scarce and exhibits sparse ground object characteristics, with uneven spatial distribution and substantial redundant information in the spectral dimension. Although 3D-CNN structures can utilize joint spatial-spectral information, how to more effectively achieve deep extraction of spatial-spectral information remains a noteworthy issue. As the core of convolutional neural networks, convolution kernels are generally regarded as information aggregators that combine spatial information and feature dimension information in local receptive fields. Convolutional neural networks consist of a series of convolutional layers, nonlinear layers, and downsampling layers, enabling them to capture image features from a global receptive field for image description. However, training a high-performance network is challenging, and much work has been done to improve network performance from the spatial dimension perspective. For example, the Residual structure achieves deep network extraction by fusing features produced by different blocks, while DenseNet enhances feature reuse through dense connections. 3D-CNN contains numerous redundant weights in feature extraction through convolution operations that simultaneously process spatial and spectral information of hyperspectral images. This redundancy is particularly prominent in joint spatial-spectral feature extraction: from the spatial dimension, ground objects in hyperspectral images are sparsely and unevenly distributed, and convolutional kernels may capture many irrelevant or low-information regions within local receptive fields; from the spectral dimension, hyperspectral data typically contains hundreds of bands with high correlation and redundancy between adjacent bands, making weight allocation of convolutional kernels along the spectral axis difficult to effectively focus on key features. Especially the redundant characteristics of spectral dimensions cause many convolutional parameters to only serve as "fillers" in high-dimensional data without fully mining deep patterns in joint spatial-spectral information. This weight redundancy not only increases computational complexity but may also weaken the model's representation capability for sparse ground objects and complex spectral features, thereby limiting 3D-CNN's performance in hyperspectral image processing. This paper designs EKGNet for hyperspectral image classification, including a context-aware mapping network that acts as an intelligent translator: accurately parsing global contextual features of input hyperspectral image patches and "translating" them into explicit instructions to guide subsequent convolution operations. These instructions act on an expert convolution system composed of K groups of pre-learned base convolutional kernels, which can be viewed as K "experts," each potentially specializing in capturing different dimensional or fundamental pattern features (such as specific spectral absorption bands, edge textures, or spatial structures). Crucially, the context-aware mapping network and expert convolution system form a tightly coupled, synergistic whole: the former generates meaningful combination weights (i.e., "expert invocation instructions") based on deep understanding of inputs, while the latter uses these real-time generated weights to dynamically "assemble" a highly adaptive expert-level convolution operation most suitable for current inputs. This ultimately achieves effective representation extraction of spatial-spectral features, which is particularly effective for sparse ground objects in hyperspectral images where spectral data is relatively scarce and belongs to the small-sample category. Feature maps obtained by different convolution kernels exhibit significant differences. Through dynamic convolution generation combined with the depth characteristics of 3D-DenseNet, features can be extracted more effectively.

\begin{figure}[h]
\centering
\includegraphics[width=0.9\linewidth]{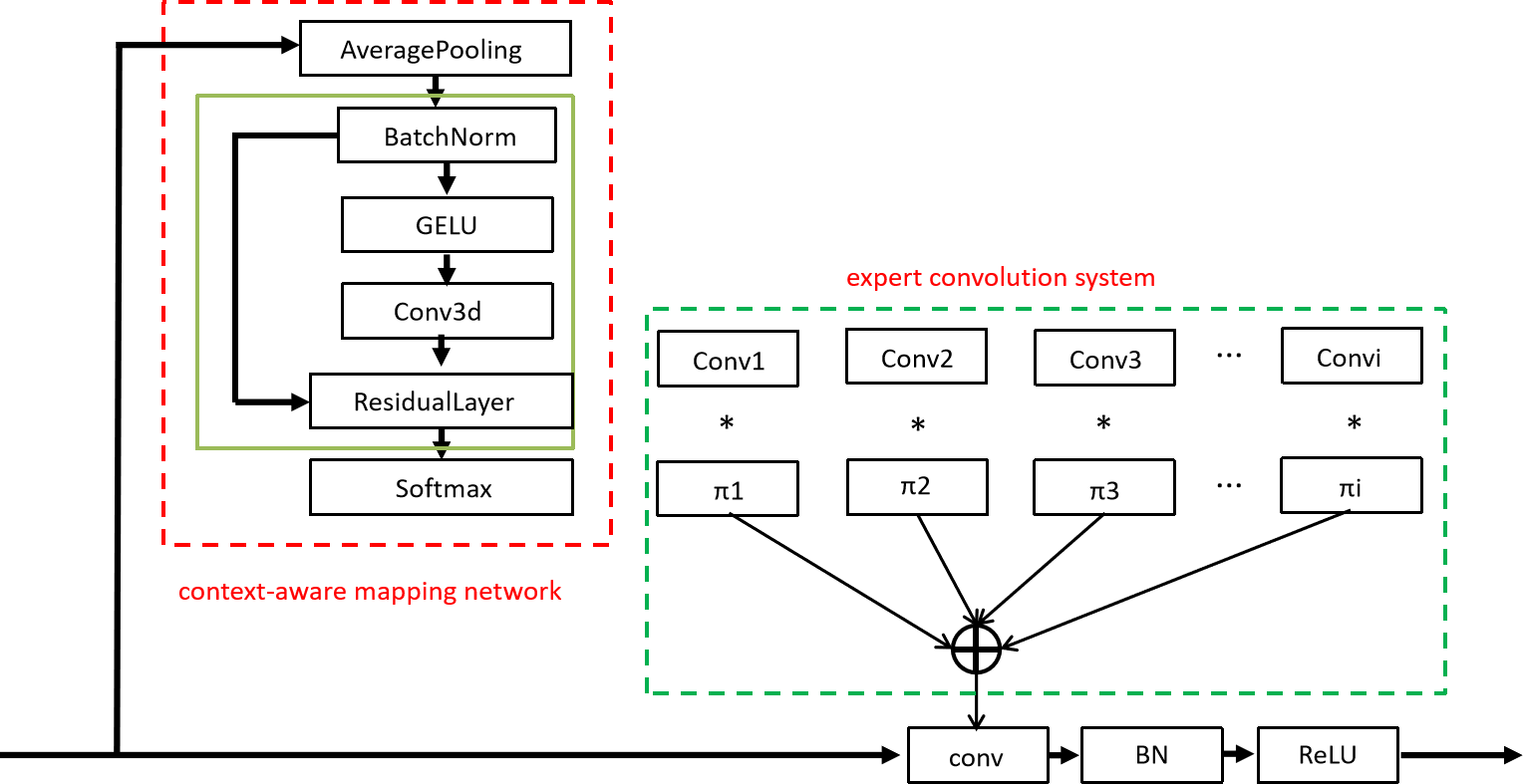}
\caption{Context-aware mapping network and dynamic expert convolution system module}
\label{fig:ekgnet_module}
\end{figure}

\subsubsection{Context-Aware Mapping Network}
The context-aware mapping network in EKGNet serves as an intelligent translator that extracts global contextual information from hyperspectral input data and generates attention weights to guide the dynamic convolution process. This module addresses the challenges of spatial-spectral feature redundancy and uneven distribution in hyperspectral images (HSI) by adaptively focusing on key patterns while suppressing irrelevant or noisy information. The network is represented as $ f_{\text{map}} $, processing input feature maps $ X \in \mathbb{R}^{B \times C \times D \times H \times W} $, where $ B $ is batch size, $ C $ is the number of input channels (spectral bands), and $ D $, $ H $, $ W $ represent depth, height, and width of spatial dimensions respectively.

The operation first extracts global context through an adaptive average pooling layer, compressing spatial-spectral dimensions into a compact representation:
$$ g = \text{AvgPool3d}(X) \in \mathbb{R}^{B \times C \times 1 \times 1 \times 1}, $$
where $ g $ captures the global context summary of the input. This pooled feature is then processed through a series of layers in the $ f_{\text{map}} $ network (implemented in the keys module). Specifically, the network consists of multiple blocks, each containing batch normalization, GELU activation, and 3D convolutional layers, followed by residual connections to enhance feature flow and stabilize training. Its structure can be described as:
$$ h_0 = g, $$
$$ h_{i+1} = \text{ResidualLayer}(\text{Conv3d}(\text{GELU}(\text{BN}(h_i))), h_i), $$
$$ \text{attn} = \text{Conv3d}(\text{BN}(h_N)) \in \mathbb{R}^{B \times K \times 1 \times 1 \times 1}, $$
where $ N $ is the number of blocks, $ K $ is the number of expert convolutional kernels, and ResidualLayer applies weighted residual connections controlled by learnable parameter $ \alpha $. The final convolution projects features to $ K $ channels, representing attention scores for each expert convolutional kernel.

These scores are converted to attention weights via softmax operation with learnable temperature parameter $ \tau $:
$$ \alpha = \text{Softmax}\left(\frac{\text{attn}}{\tau}\right) \in \mathbb{R}^{B \times K}, $$
where $ \alpha = \{\alpha_1, \dots, \alpha_K\} $ represents combination weights for $ K $ expert convolutional kernels, and $ \tau $ controls sharpness of attention distribution. Temperature is dynamically adjusted during training through update temperature method to ensure stable convergence, gradually decreasing from initial value to focus on dominant convolutional kernels.

This context mapping process ensures attention weights $ \alpha $ are customized according to specific inputs, enabling the model to prioritize relevant spatial-spectral patterns (e.g., specific absorption bands or texture edges) while mitigating impact of redundant or noisy dimensions. By integrating global context through pooling and lightweight yet expressive networks, the context-aware mapping network efficiently translates input characteristics into actionable instructions for the expert convolution system.

\subsubsection{Dynamic Kernel Generation Expert Convolution System}
The expert convolution system in EKGNet utilizes attention weights $ \alpha $ generated by the context-aware mapping network to dynamically construct a customized convolutional kernel for each input sample. This system addresses limitations of static 3D convolution, where fixed receptive fields often aggregate irrelevant spatial-spectral features, leading to overfitting and poor generalization in HSI classification. By employing a set of $ K $ learnable base convolutional kernels (viewed as "experts" focusing on different feature patterns), the system generates an adaptive convolution kernel $ W_{\text{dyn}} $ that optimizes feature extraction for given inputs.

Formally, the system maintains $ K $ base convolutional kernels $ \{W_1, W_2, \dots, W_K\} $, where each $ W_k \in \mathbb{R}^{C_{\text{out}} \times (C_{\text{in}}/G) \times S \times S \times S} $, with $ C_{\text{out}} $ output channels, $ C_{\text{in}} $ input channels, $ G $ groups, and $ S $ kernel size. These kernels are stored as learnable parameters weight and initialized via truncated normal distribution to ensure training stability.

The dynamic convolution kernel $W_{\text{dyn}}$ is computed using attention weights $\alpha$ through weighted combination of base kernels:
$$ W_{\text{dyn}} = \sum_{k=1}^K \alpha_k W_k, $$
where $W_{\text{dyn}} \in \mathbb{R}^{C_{\text{out}} \times (C_{\text{in}}/G) \times S \times S \times S}$. Similarly, if bias is enabled, dynamic bias term is computed:
$$ b_{\text{dyn}} = \sum_{k=1}^K \alpha_k b_k, $$
where $b_k \in \mathbb{R}^{C_{\text{out}}}$ is learnable bias for each expert kernel. This aggregation is efficiently performed through matrix multiplication:
\begin{align*}
\text{aggregate\_weight} &= \text{mm}(\alpha, \text{weight.view}(K, -1)).\text{view}(B \cdot C_{\text{out}}, C_{\text{in}}/G, S, S, S), \\
\text{aggregate\_bias} &= \text{mm}(\alpha, b).\text{view}(-1).
\end{align*}

The generated $W_{\text{dyn}}$ is applied to input $X$ via 3D convolution operation:
$$ Y = \text{Conv3d}(X, W_{\text{dyn}}, b_{\text{dyn}}, \text{stride}, \text{padding}, \text{dilation}, G \cdot B), $$
where $Y \in \mathbb{R}^{B \times C_{\text{out}} \times D' \times H' \times W'}$ is output feature map, and grouped convolution considers batch processing to ensure each sample uses its unique dynamic convolution kernel. Convolution parameters (stride, padding, dilation) are configurable.

This dynamic kernel generation enables EKGNet to adaptively adjust receptive fields and feature focus based on spatial-spectral characteristics of inputs. For instance, in regions with sparse ground objects, the system may emphasize kernels capturing fine-grained spectral details, while in areas with complex spatial patterns, it may prioritize kernels sensitive to structural features. By integrating this mechanism into the 3D-DenseNet framework, EKGNet achieves enhanced feature reuse and representation capability without introducing excessive parameters. The synergy between context-aware mapping network and expert convolution system ensures EKGNet effectively mitigates challenges posed by high-dimensional redundancy and sparse ground object distribution, achieving exceptional classification performance.

\begin{figure}[h]
\centering
\includegraphics[width=0.9\linewidth]{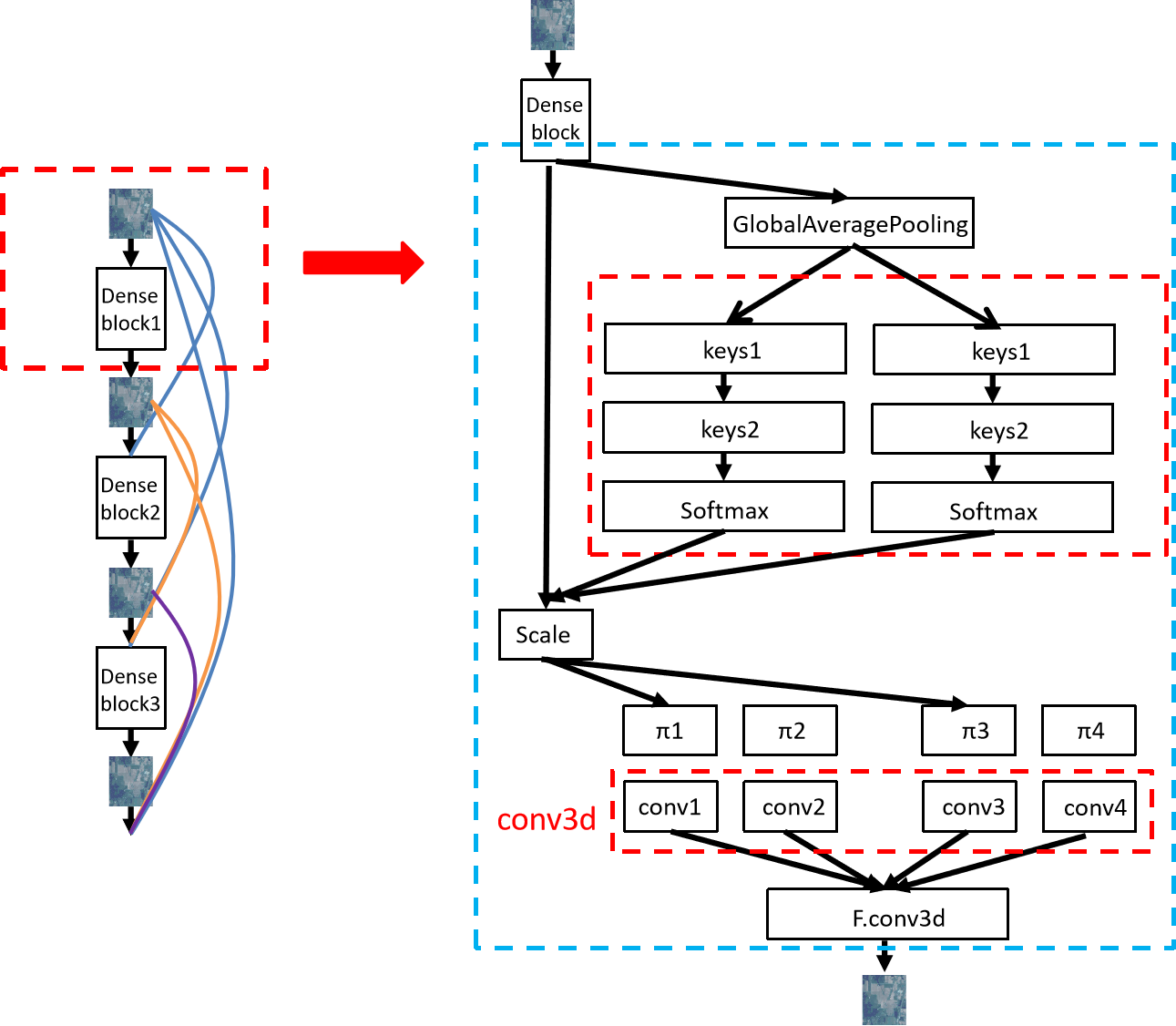}
\caption{EKGNet design in 3D-DenseNet's dense block}
\label{fig:dense_block}
\end{figure}

\subsection{3D-CNN Framework for Hyperspectral Image Feature Extraction}

We introduce two key modifications to the original 3D-DenseNet architecture to enhance simplicity and computational efficiency.

\subsubsection{Exponentially Increasing Growth Rate}
The original DenseNet design adds $ k $ new feature maps per layer, where $ k $ is a constant growth rate. As shown in \cite{huang2017densely}, deeper layers in DenseNet tend to rely more on high-level features than low-level ones, which motivated our improvement through strengthened short connections. We achieve this by progressively increasing the growth rate with depth, enhancing the proportion of features from later layers relative to earlier ones. For simplicity, we set the growth rate as $ k=2^{m-1}k_0 $, where $ m $ is the dense block index and $ k_0 $ is a constant. This growth rate configuration introduces no additional hyperparameters. The "increasing growth rate" strategy places a larger proportion of parameters in the model's later layers, significantly improving computational efficiency while potentially reducing parameter efficiency in some cases. Depending on specific hardware constraints, trading one for the other may be advantageous \cite{liu2017learning}.

\begin{figure}[h]
\centering
\includegraphics[width=0.9\linewidth]{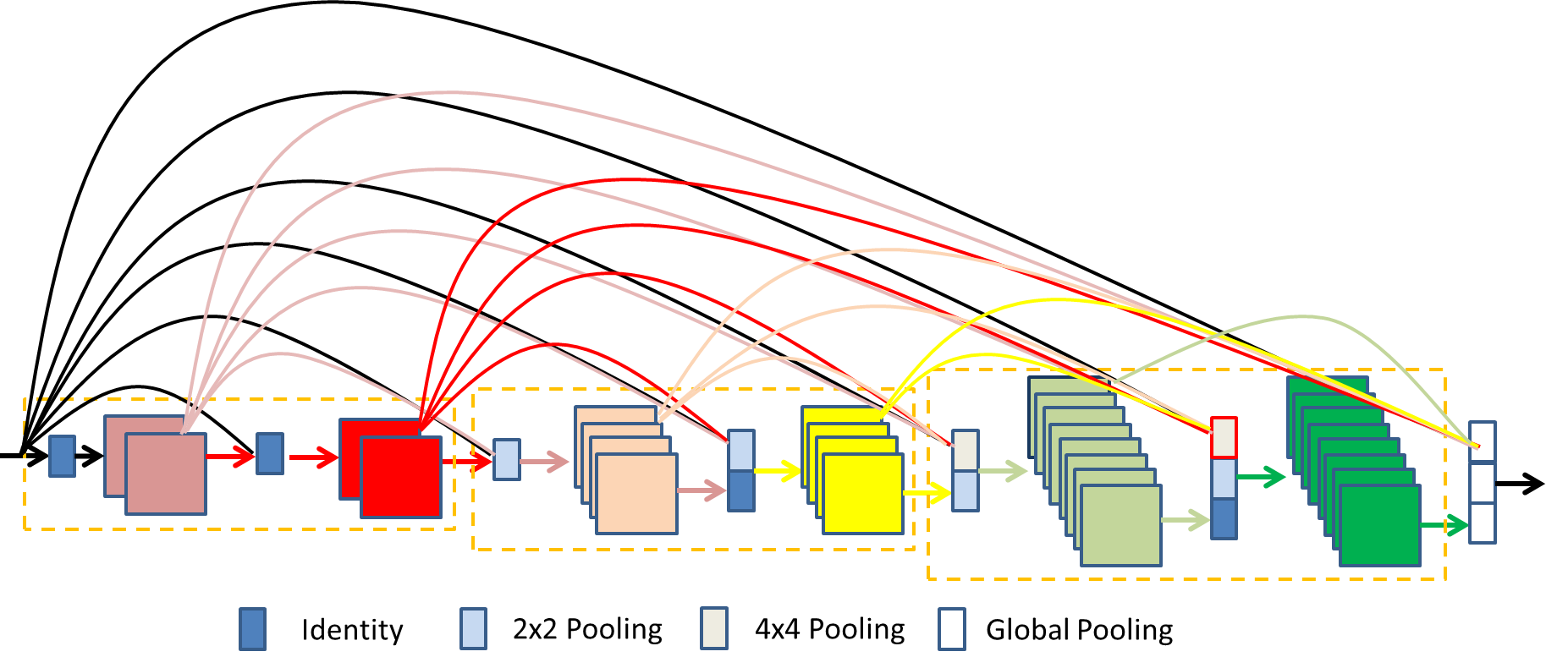}
\caption{Proposed DenseNet variant with two key differences from original DenseNet: (1) Direct connections between layers with different feature resolutions; (2) Growth rate doubles when feature map size reduces (third yellow dense block generates significantly more features than the first block)}
\label{fig:densenet_variant}
\end{figure}

\subsubsection{Fully Dense Connectivity}
To encourage greater feature reuse than the original DenseNet architecture, we connect the input layer to all subsequent layers across different dense blocks (see Figure~\ref{fig:densenet_variant}). Since dense blocks have different feature resolutions, we downsample higher-resolution feature maps using average pooling when connecting them to lower-resolution layers.

\begin{figure}[t]
\centering
\includegraphics[width=0.9\linewidth]{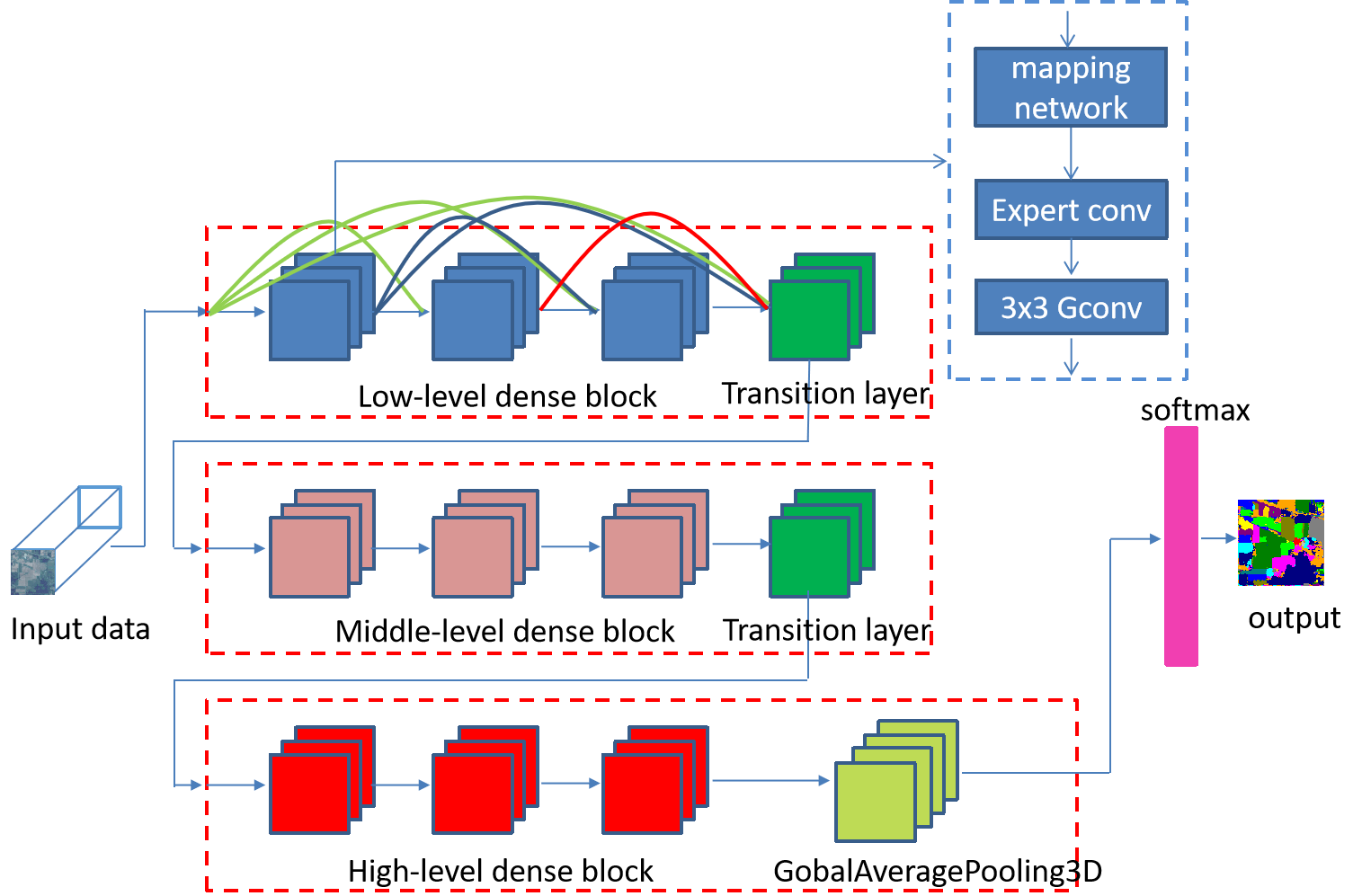}
\caption{Overall architecture of our EKGNet, incorporating the modified 3D-DenseNet framework}
\label{fig:ekgnet_arch}
\end{figure}

\section{Experiments and Analysis}

To evaluate the performance of EKGNet, we conducted experiments on three representative hyperspectral datasets: Indian Pines, Pavia University, and Kennedy Space Center (KSC). The classification metrics include Overall Accuracy (OA), Average Accuracy (AA), and Kappa coefficient.

\subsection{Datasets}
\subsubsection{Indian Pines Dataset}
\begin{figure}[h]
\centering
\includegraphics[width=0.9\linewidth]{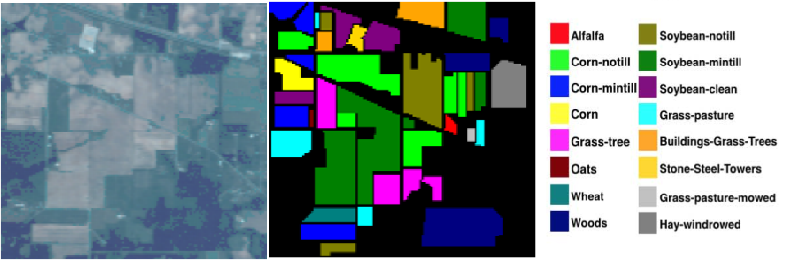}
\caption{False color composite and ground truth labels of Indian Pines dataset}
\label{fig:indian_pines}
\end{figure}

The Indian Pines dataset was collected in June 1992 by the AVIRIS (Airborne Visible/Infrared Imaging Spectrometer) sensor over a pine forest test site in northwestern Indiana, USA. The dataset consists of 145$\times$145 pixel images with a spatial resolution of 20 meters, containing 220 spectral bands covering the wavelength range of 0.4--2.5$\mu$m. In our experiments, we excluded 20 bands affected by water vapor absorption and low signal-to-noise ratio (SNR), utilizing the remaining 200 bands for analysis. The dataset encompasses 16 land cover categories including grasslands, buildings, and various crop types. Figure~\ref{fig:indian_pines} displays the false-color composite image and spatial distribution of ground truth samples.

\subsubsection{Pavia University Dataset}
\begin{figure}[h]
\centering
\includegraphics[width=0.9\linewidth]{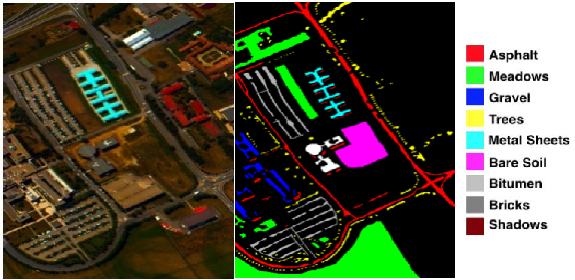}
\caption{False color composite and ground truth labels of Pavia University dataset}
\label{fig:pavia_university}
\end{figure}

The Pavia University dataset was acquired in 2001 by the ROSIS imaging spectrometer over the Pavia region in northern Italy. The dataset contains images of size 610$\times$340 pixels with a spatial resolution of 1.3 meters, comprising 115 spectral bands in the wavelength range of 0.43--0.86$\mu$m. For our experiments, we removed 12 bands containing strong noise and water vapor absorption, retaining 103 bands for analysis. The dataset includes 9 land cover categories such as roads, trees, and roofs. Figure~\ref{fig:pavia_university} shows the spatial distribution of different classes.

\subsubsection{Kennedy Space Center Dataset}
The KSC dataset was collected on March 23, 1996 by the AVIRIS imaging spectrometer over the Kennedy Space Center in Florida. The AVIRIS sensor captured 224 spectral bands with 10 nm width, centered at wavelengths from 400 to 2500 nm. Acquired from an altitude of approximately 20 km, the dataset has a spatial resolution of 18 meters. After removing bands affected by water absorption and low signal-to-noise ratio (SNR), we used 176 bands for analysis, which represent 13 defined land cover categories.

\subsection{Experimental Analysis}
EKGNet was trained for 80 epochs on all three datasets using the Adam optimizer. The experiments were conducted on a platform with four 80GB A100 GPUs. For analysis, we employed the ekgnet-base architecture with three stages, where each stage contained 4, 6, and 8 dense blocks respectively. The growth rates were set to 8, 16, and 32, with 4 heads. The 3×3 group convolution used 4 groups, gate factor was 0.25, and the compression ratio was 16.

\subsubsection{Data Partitioning Ratio}
For hyperspectral data with limited samples, the training set ratio significantly impacts model performance. To systematically evaluate the sensitivity of data partitioning strategies, we compared the model's generalization performance under different Train/Validation/Test ratios. Experiments show that with limited training samples, a 6:1:3 ratio effectively balances learning capability and evaluation reliability on Indian Pines - this configuration allocates 60\% samples for training, 10\% for validation (enabling early stopping to prevent overfitting), and 30\% for statistically significant testing. EKGNet adopted 5:1:4 ratio on Pavia University and KSC datasets, with 11×11 neighboring pixel blocks to balance local feature extraction and spatial context integrity.

\begin{table}[h]
\begin{minipage}{\textwidth}
\centering
\makeatletter
\def\@makecaption#1#2{%
    \vskip\abovecaptionskip
    \centering 
    \small #1: #2\par
    \vskip\belowcaptionskip
}
\makeatother
\caption{OA, AA and Kappa metrics for different training set ratios on the Indian Pines dataset}
\begin{adjustbox}{width=0.5\columnwidth}
\begin{tabular}{cccc}
\toprule
Training Ratio & OA & AA & Kappa \\
\midrule
2:1:7 & 89.04 & 72.78 & 87.46 \\
3:1:6 & 94.89 & 88.35 & 94.18 \\
4:1:5 & 98.40 & 97.71 & 98.18 \\
5:1:4 & 98.76 & 96.95 & 98.58 \\
6:1:3 & 99.09 & 99.26 & 98.96 \\
\bottomrule
\end{tabular}
\label{tab:indian_ratios}
\end{adjustbox}
    \end{minipage}

 \vspace*{10pt} 
 
\begin{minipage}{\textwidth}
\centering
\makeatletter
\def\@makecaption#1#2{%
    \vskip\abovecaptionskip
    \centering 
    \small #1: #2\par
    \vskip\belowcaptionskip
}
\makeatother
\caption{OA, AA and Kappa metrics for different training set ratios on the Pavia University dataset}
\begin{tabular}{cccc}
\toprule
Training Ratio & OA & AA & Kappa \\
\midrule
2:1:7 & 99.27 & 99.09 & 99.03 \\
3:1:6 & 99.33 & 99.13 & 99.12 \\
4:1:5 & 99.54 & 99.54 & 99.39 \\
5:1:4 & 99.89 & 99.80 & 99.86 \\
6:1:3 & 99.86 & 99.74 & 99.82 \\
\bottomrule
\end{tabular}
\label{tab:pavia_ratios}
        \end{minipage}

 \vspace*{10pt} 
 
\begin{minipage}{\textwidth}        
\centering
\makeatletter
\def\@makecaption#1#2{%
    \vskip\abovecaptionskip
    \centering 
    \small #1: #2\par
    \vskip\belowcaptionskip
}
\makeatother
\caption{OA, AA and Kappa metrics for different training set ratios on the KSC dataset}
\begin{tabular}{cccc}
\toprule
Training Ratio & OA & AA & Kappa \\
\midrule
2:1:7 & 96.32 & 94.22 & 95.90 \\
3:1:6 & 98.62 & 97.94 & 98.47 \\
4:1:5 & 98.81 & 98.54 & 98.67 \\
5:1:4 & 99.28 & 99.17 & 99.20 \\
6:1:3 & 99.23 & 99.11 & 99.15 \\
\bottomrule
\end{tabular}
\label{tab:ksc_ratios}
        \end{minipage}
\end{table}

\subsubsection{Neighboring Pixel Blocks}
The network pads the input 145×145×103 image (using Indian Pines as example) to 155×155×103, then extracts M×N×L neighboring blocks where M×N is spatial size and L is spectral dimension. Large original images hinder feature extraction, slow processing, and increase memory demands. Thus neighboring block processing is adopted, where block size is a crucial hyperparameter. However, blocks cannot be too small as this limits receptive fields. As shown in Tables \ref{tab:indian_blocks}-\ref{tab:ksc_blocks}, accuracy improves significantly from size 7 to 15 on Indian Pines, with similar trends on Pavia University and KSC datasets. We selected size 15 for all datasets.

\begin{table*}[!ht]
\begin{minipage}{\textwidth}
\centering
\makeatletter
\def\@makecaption#1#2{%
    \vskip\abovecaptionskip
    \centering 
    \small #1: #2\par
    \vskip\belowcaptionskip
}
\makeatother
\caption{OA, AA and Kappa metrics for different block sizes on Indian Pines}
\begin{tabular}{cccc}
\toprule
Block Size (M=N) & OA & AA & Kappa \\
\midrule
7 & 95.67 & 95.52 & 95.06 \\
9 & 97.56 & 97.91 & 97.22 \\
11 & 99.09 & 99.26 & 98.96 \\
13 & 99.19 & 99.32 & 99.07 \\
15 & 99.84 & 99.51 & 99.81 \\
17 & 99.77 & 98.73 & 99.74 \\
\bottomrule
\end{tabular}
\label{tab:indian_blocks}
        \end{minipage}

 \vspace*{10pt} 

\begin{minipage}{\textwidth}
\centering
\caption{OA, AA and Kappa metrics for different block sizes on Pavia University}
\begin{tabular}{cccc}
\toprule
Block Size (M=N) & OA & AA & Kappa \\
\midrule
7 & 98.56 & 98.50 & 98.09 \\
9 & 99.70 & 99.48 & 99.60 \\
11 & 99.89 & 99.80 & 99.86 \\
13 & 99.94 & 99.88 & 99.92 \\
15 & 99.98 & 99.99 & 99.98 \\
17 & 99.26 & 98.99 & 99.02 \\
\bottomrule
\end{tabular}
\label{tab:pavia_blocks}
\end{minipage}

 \vspace*{10pt} 
 
\begin{minipage}{\textwidth}
\centering
\makeatletter
\def\@makecaption#1#2{%
    \vskip\abovecaptionskip
    \centering 
    \small #1: #2\par
    \vskip\belowcaptionskip
}
\makeatother
\caption{OA, AA and Kappa metrics for different block sizes on KSC}
\begin{tabular}{cccc}
\toprule
Block Size (M=N) & OA & AA & Kappa \\
\midrule
7 & 94.57 & 92.68 & 93.95 \\
9 & 97.55 & 96.36 & 97.27 \\
11 & 99.28 & 99.17 & 99.20 \\
13 & 99.42 & 99.22 & 99.36 \\
15 & 99.95 & 99.92 & 99.95 \\
17 & 99.86 & 99.77 & 99.84 \\
\bottomrule
\end{tabular}
\label{tab:ksc_blocks}
        \end{minipage}
\end{table*}

\subsection{Network Parameters}
We divided EKGNet into base and large variants. Table \ref{tab:params} shows the OA, AA, and Kappa metrics tested on the Indian Pines dataset. The large model has higher parameters, and as the parameter count increases, the network accuracy improves accordingly. However, compared to the rapid parameter growth in dense blocks, the accuracy improvement is not substantial. Therefore, we believe EKGNet-base already possesses excellent generalization and classification capabilities.

\begin{table*}[!ht]
\centering
\makeatletter
\def\@makecaption#1#2{%
    \vskip\abovecaptionskip
    \centering 
    \small #1: #2\par
    \vskip\belowcaptionskip
}
\makeatother
\caption{EKGNet model configurations and performance metrics}
\begin{tabular}{cccccc}
\toprule
Model & Stages/Dense Block & Growth Rate & OA & AA & Kappa \\
\midrule
EKGNet-base & 4,6,8 & 8,16,32 & 99.84 & 99.51 & 99.81 \\
EKGNet-large & 14,14,14 & 8,16,32 & 99.89 & 99.72 & 99.83 \\
\bottomrule
\end{tabular}
\label{tab:params}
\end{table*}

\subsection{Experimental Results and Analysis}
On the Indian Pines dataset, EKGNet's input size is 17$\times$17$\times$200; on Pavia University, it's 17$\times$17$\times$103; and on KSC, it's 17$\times$17$\times$176. We compared EKGNet-base with SSRN, 3D-CNN, 3D-SE-DenseNet, SpectralFormer, LGCNET, and DGCNET, as shown in Tables \ref{tab:indian_results} and \ref{tab:pavia_results}. EKGNet's three variants achieved leading accuracy overall. Figure \ref{fig:training} shows the loss and accuracy changes during training and validation, demonstrating rapid convergence of the loss function and stable accuracy improvement.

\begin{table*}[!ht]
\centering
\makeatletter
\def\@makecaption#1#2{%
    \vskip\abovecaptionskip
    \centering 
    \small #1: #2\par
    \vskip\belowcaptionskip
}
\makeatother
\caption{Classification accuracy comparison (\%) on Indian Pines dataset}
\label{tab:indian_results}
\resizebox{\textwidth}{!}{
\begin{tabular}{@{}lccccccc@{}}
\toprule
Class & SSRN & 3D-CNN & 3D-SE-DenseNet & Spectralformer & LGCNet & DGCNet & EKGNet \\
\midrule
1 & 100 & 96.88 & 95.87 & 70.52 & 100 & 100 & 100 \\
2 & 99.85 & 98.02 & 98.82 & 81.89 & 99.92 & 99.47 & 100 \\
3 & 99.83 & 97.74 & 99.12 & 91.30 & 99.87 & 99.51 & 100 \\
4 & 100 & 96.89 & 94.83 & 95.53 & 100 & 97.65 & 100 \\
5 & 99.78 & 99.12 & 99.86 & 85.51 & 100 & 100 & 100 \\
6 & 99.81 & 99.41 & 99.33 & 99.32 & 99.56 & 99.88 & 100 \\
7 & 100 & 88.89 & 97.37 & 81.81 & 95.83 & 100 & 100 \\
8 & 100 & 100 & 100 & 75.48 & 100 & 100 & 100 \\
9 & 0 & 100 & 100 & 73.76 & 100 & 100 & 100 \\
10 & 100 & 100 & 99.48 & 98.77 & 99.78 & 98.85 & 99.67 \\
11 & 99.62 & 99.33 & 98.95 & 93.17 & 99.82 & 99.72 & 99.87 \\
12 & 99.17 & 97.67 & 95.75 & 78.48 & 100 & 99.56 & 99.44 \\
13 & 100 & 99.64 & 99.28 & 100 & 100 & 100 & 100 \\
14 & 98.87 & 99.65 & 99.55 & 79.49 & 100 & 99.87 & 100 \\
15 & 100 & 96.34 & 98.70 & 100 & 100 & 100 & 100 \\
16 & 98.51 & 97.92 & 96.51 & 100 & 97.73 & 98.30 & 93.10 \\
\midrule
OA & 99.62$\pm$0.00 & 98.23$\pm$0.12 & 98.84$\pm$0.18 & 81.76 & 99.85$\pm$0.04 & 99.58 & 99.84 \\
AA & 93.46$\pm$0.50 & 98.80$\pm$0.11 & 98.42$\pm$0.56 & 87.81 & 99.53$\pm$0.23 & 99.55 & 99.51 \\
$\kappa$ & 99.57$\pm$0.00 & 97.96$\pm$0.53 & 98.60$\pm$0.16 & 79.19 & 99.83$\pm$0.05 & 99.53 & 99.81 \\
\bottomrule
\end{tabular}
}
\end{table*}

\begin{figure*}[!ht]
\centering
\includegraphics[width=0.8\linewidth]{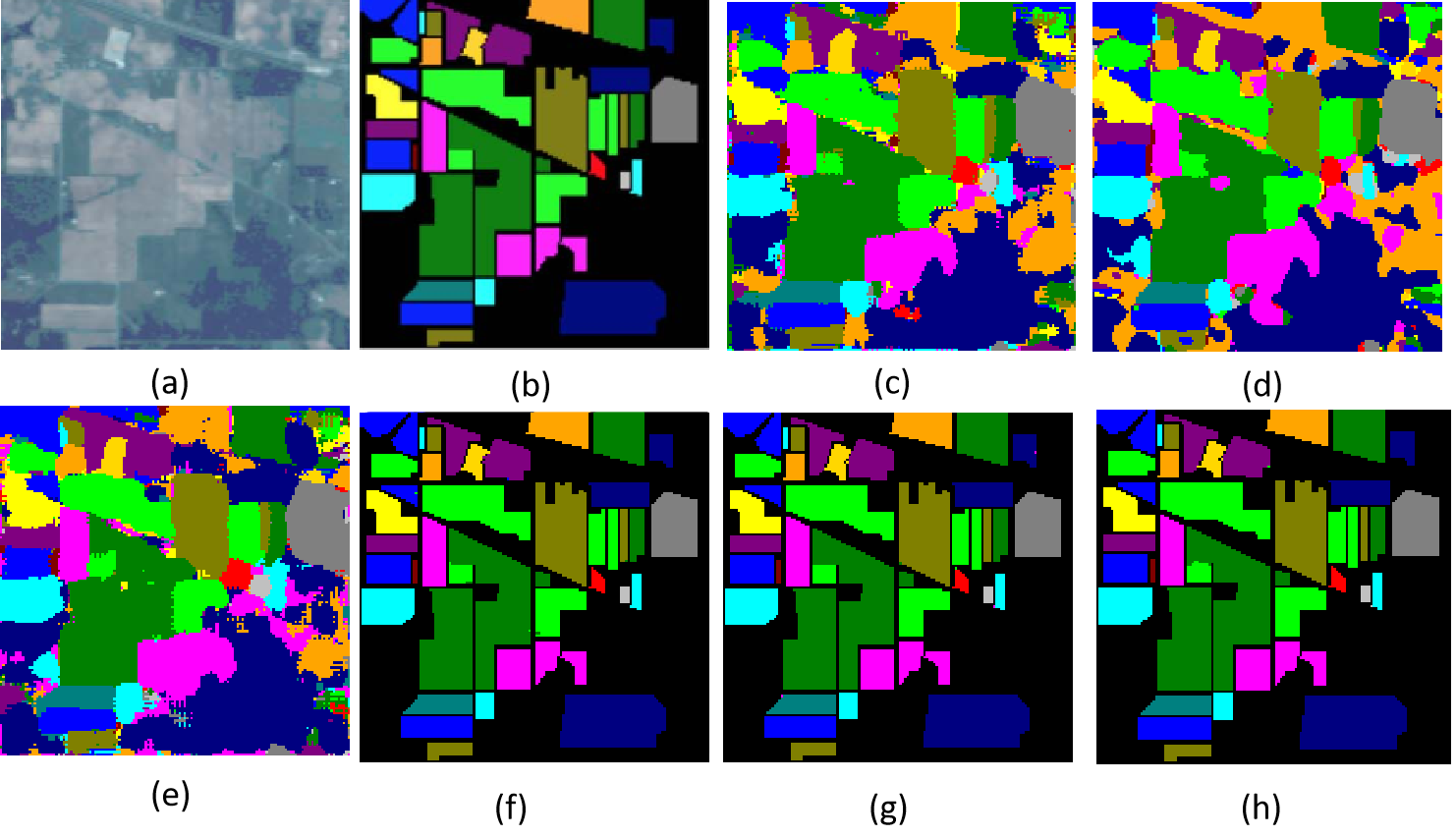}
\caption{Classification results comparison for Indian Pines dataset: (a) False color image, (b) Ground-truth labels, (c)-(i) Classification results of SSRN, 3D-CNN, 3D-SE-DenseNet, LGCNet, DGCNet, and EKGNet}
\label{fig:classification}
\end{figure*}

\begin{figure*}[ht]
\centering
\includegraphics[width=0.8\linewidth]{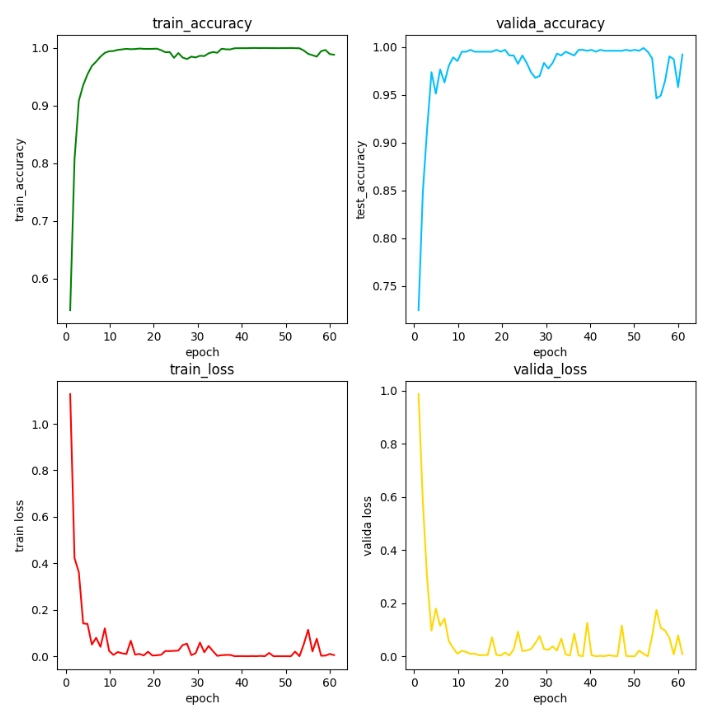}
\caption{Training and validation curves of EKGNet showing loss and accuracy evolution}
\label{fig:training_curve}
\end{figure*}

\begin{table*}[!ht]
\centering
\makeatletter
\def\@makecaption#1#2{%
    \vskip\abovecaptionskip
    \centering 
    \small #1: #2\par
    \vskip\belowcaptionskip
}
\makeatother
\caption{Classification accuracy (\%) comparison on Pavia University dataset}
\label{tab:pavia_results}
\resizebox{\textwidth}{!}{
\begin{tabular}{@{}lcccccc@{}}
\toprule
Class & SSRN & 3D-CNN & 3D-SE-DenseNet & Spectralformer & LGCNet & EKGNet \\
\midrule
1 & 89.93 & 99.96 & 99.32 & 82.73 & 100 & 99.88 \\
2 & 86.48 & 99.99 & 99.87 & 94.03 & 100 & 100 \\
3 & 99.95 & 99.64 & 96.76 & 73.66 & 99.88 & 100 \\
4 & 95.78 & 99.83 & 99.23 & 93.75 & 100 & 100 \\
5 & 97.69 & 99.81 & 99.64 & 99.28 & 100 & 100 \\
6 & 95.44 & 99.98 & 99.80 & 90.75 & 100 & 100 \\
7 & 84.40 & 97.97 & 99.47 & 87.56 & 100 & 100 \\
8 & 100 & 99.56 & 99.32 & 95.81 & 100 & 100 \\
9 & 87.24 & 100 & 100 & 94.21 & 100 & 100 \\
\midrule
OA & 92.99$\pm$0.39 & 99.79$\pm$0.01 & 99.48$\pm$0.02 & 91.07 & 99.99$\pm$0.00 & 99.98 \\
AA & 87.21$\pm$0.25 & 99.75$\pm$0.15 & 99.16$\pm$0.37 & 90.20 & 99.99$\pm$0.01 & 99.99 \\
$\kappa$ & 90.58$\pm$0.18 & 99.87$\pm$0.27 & 99.31$\pm$0.03 & 88.05 & 99.99$\pm$0.00 & 99.98 \\
\bottomrule
\end{tabular}
}
\end{table*}

\section{Conclusion}
This paper introduces two modules on 3D convolution kernels: a context-associated mapping network and a dynamic kernel generation mechanism. The context-associated mapping module translates the global hyperspectral context information of the input into instructions for combining basic convolution kernels, while the dynamic convolution kernels are composed of K groups of basic kernels, akin to K distinct experts specializing in addressing feature patterns of different dimensions. The mapping module and the dynamic kernel generation mechanism form a tightly coupled system, where the former generates meaningful combination weights based on the input, and the latter utilizes these weights to construct a highly adaptive expert convolution system. This approach further enhances the model's ability to represent spatio-spectral joint information. The dynamic design enables EGKNet to flexibly adapt to the feature demands of different spatial regions without relying on a single static convolution kernel, while effectively skipping redundant information and reducing computational complexity. EGKNet leverages the 3D-DenseNet architecture to extract spatial structure and spectral information of more critical features, providing an efficient solution for hyperspectral image classification. It successfully addresses challenges posed by sparse ground object distribution and spectral redundancy.

\section*{Data Availability Statement}
The datasets used in this study are publicly available and widely used benchmark datasets in the hyperspectral image analysis community.

{\small
\bibliographystyle{template}
\bibliography{template}
}

\end{document}